\def\year{2021}\relax
\documentclass[letterpaper]{article} 
\usepackage{aaai21}  
\usepackage{times}  
\usepackage{helvet} 
\usepackage{courier}  
\usepackage[hyphens]{url}  
\usepackage{graphicx} 
\usepackage{natbib}  
\usepackage{caption} 
\usepackage{mathtools}
\usepackage{bm}
\usepackage{multirow}
\usepackage{mathbbol}
\usepackage{subfigure}
\usepackage{booktabs}

\title{A General Offline Reinforcement Learning Framework for Interactive Recommendation}
\author{
    Teng Xiao, Donglin Wang\footnote{Corresponding author.}\\}
\affiliations{
    Machine Intelligence Lab (MiLAB), AI Division, School of Engineering, Westlake University\\
    tengxiao01@gmail.com, wangdonglin@westlake.edu.cn
}

\begin{document}
\maketitle
\begin{abstract}
This paper studies the problem of learning interactive recommender systems from logged  feedbacks without any exploration in online environments. We address the problem by proposing a general offline reinforcement learning framework for recommendation, which enables maximizing cumulative user rewards without online exploration. Specifically, we first introduce a probabilistic generative model for  interactive recommendation, and then propose an effective inference algorithm for discrete and stochastic policy learning based on logged  feedbacks. In order to perform offline learning more effectively, we propose five approaches to minimize the distribution mismatch between the logging policy and recommendation policy: support constraints, supervised regularization, policy constraints, dual constraints and reward extrapolation. We conduct extensive  experiments on two public real-world datasets,  demonstrating that the proposed methods can achieve superior performance over existing supervised learning and reinforcement learning methods for  recommendation.
\end{abstract}

\section{Introduction}
Reinforcement learning (RL)   is a powerful paradigm for interactive or sequential recommender systems (RS), since it  can maximize users' long-term satisfaction with the system and constantly adapt to users’ shifting interest (state). However, training optimal RL algorithms  requires  large amounts of interactions with   environments~\cite{kakade2003sample}, which is impractical in the recommendation context. Performing online interaction (on-policy) learning  would hurt  users’ experiences and the revenue of the platform. Since logged feedbacks of users are often abundant and cheap, an alternative is to make use of them and to learn a near-optimal recommendation policy offline  before we deploy it online. 

\noindent Although there are some offline (off-policy) RL algorithms have been proposed in the continuous robot control domain~\cite{fujimoto2019off,kumar2019stabilizing}, the problem that how to build an effective offline RL framework for recommendation  involving  large numbers of discrete actions and logged feedbacks remains an open one. Learning an effective  recommendation policy  from logged  feedbacks faces the following challenges simultaneously:\\
 \textbf{(a)} Discrete stochastic policy. In the testing, instead of  recommending only one item, the RS  requires generating a top-k item list according a discrete policy. Training a deterministic policy violates the simple machine learning principle: test and train conditions must match (see \S\ref{Sec:4} for details). \\
\textbf{(b)} 
Extrapolation error. Extrapolation error is an error in off-policy value learning
which is introduced by the mismatch between the dataset and true state-action visitation of the current policy~\cite{fujimoto2019off,siegel2019keep}. This problem is even more serious for the recommendation involving large numbers of discrete actions. \\
\textbf{(c)} Unknown logging policy. The feedbacks  typically come from a unknown mixture of previous   policies. The method that requires estimating logging policy  are limited by their ability to accurately estimate the unknown logging policy.

\noindent Recent effort~\cite{DBLP:conf/wsdm/ChenBCJBC19} for off-policy recommendation tries to alleviate the problem (a) by utilizing the inverse propensity score (IPS)  with model-free policy gradient algorithm. However the IPS method suffers from high variance~\cite{DBLP:journals/jmlr/SwaminathanJ15} and will still be biased if  the logging policy does not span the support of the optimal policy~\cite{DBLP:conf/kdd/SachdevaSJ20,DBLP:conf/uai/LiuSAB19}. Moreover, it  cannot  address other  problems mentioned above, i.e., (b) and (c).  Prior works~\cite{DBLP:conf/www/ZhengZZXY0L18,DBLP:conf/kdd/ZhaoZDXTY18} try to build RL-based recommendation algorithms by directly utilizing the vanilla  Q-learning~\cite{sutton2018reinforcement}, an off-policy RL algorithm. However the Q-learning is a deterministic policy method and also suffers from the extrapolation error due to no interaction with online environments~\cite{fujimoto2019off}. 


\noindent To address the aforementioned defects, in this paper,  we propose a general and effective offline  learning framework for interactive RS. We first formalize the interactive recommendation as a probabilisitic inference problem, and then propose a discrete stochastic actor-critic algorithm to maximize cumulative rewards based on the probabilistic formulation. In order to
reduce the extrapolation error, we propose five regularization techniques: support constraints, supervised regularization, policy constraints, dual constraints and reward extrapolation for offline learning, which can constrain the mismatch between the recommendation policy and the unknown logging policy. Our approaches  can be viewed as a combination of supervised learning and off-policy reinforcement learning for recommendation with discrete actions.  We show that such combination is critical for improving the performance of  recommendation  in the offline  setting. We highlight that we are the first  to systemically study the offline learning problem in the interactive recommendation context. Our contributions can be summarized as:\\
(1) We propose a discrete stochastic RL algorithm to maximize  cumulative rewards for interactive recommendation. \\
(2) We propose a general offline learning framework  for interactive recommendation with logged feedbacks, including support constraints, supervised regularization, policy constraints, dual constraints and reward extrapolation. \\
(3) We conduct extensive offline experiments on two real-world public datasets,  empirically demonstrating the proposed methods can achieve superior performance over existing learning methods for recommendation.

\section{Related Work}
\subsection{Offline Learning for Recommendation} 
 A line of work  closely related to ours is batch bandit learning~\cite{DBLP:conf/nips/SwaminathanJ15,DBLP:journals/jmlr/SwaminathanJ15,DBLP:conf/iclr/JoachimsSR18,DBLP:conf/kdd/SachdevaSJ20,saito2020unbiased}, which mainly utilizes IPS  to correct the selection bias from the logging policy and can be viewed as a special case of RL with one-step decision making. However, interactive recommendations typically have an effect on user behavior. Thus,  we focus on the full RL setting in this paper, where the user state depends on past actions.  Recently, \citeauthor{DBLP:conf/wsdm/ChenBCJBC19} (\citeyear{DBLP:conf/wsdm/ChenBCJBC19}) apply off-policy correction to address selection biases in logged feedbacks.
The key problem of this method is that the IPS  suffers huge variance which grows exponentially with the horizon~\cite{DBLP:conf/wsdm/ChenBCJBC19}. An  alternative to offline learning in RS is the use of model-based RL algorithms. Model-based  methods~\cite{DBLP:conf/kdd/ZouXDS0Y19,DBLP:conf/wsdm/ZouXDZB0NY20,DBLP:conf/icml/Chen0LJQS19,DBLP:conf/nips/BaiGW19}  require training a model using the log data to simulate  the user environment and assist the policy learning.  However, these methods heavily depend on the accuracy of the trained  model and are computationally more complex than model-free methods. More recently,  several  methods~\cite{DBLP:conf/kdd/WangZHZ18,DBLP:conf/sigir/XinKAJ20,DBLP:conf/kdd/GongZDLGSOZ19} combine reinforcement learning and/or auxiliary task learning to improve recommendation performance.  A simple combination of several distinct losses helps  learning from logged feedbacks; however, it is more appealing to have a
single principled loss and a general framework that is applicable to offline learning from logged  feedbacks.  In this work, we focus on providing  a general offline learning framework to maximize cumulative rewards for interactive  recommendation.

\subsection{Interactive Recommendation} To address  interactive  or sequential recommendation problem, early methods~\cite{rendle2010factorizing,xiao2019dynamic2} utilize Markov Chain to model sequential patterns of users. Since  these methods fail to model complicated relations between interactions, a number of Recurrent Neural Network (RNN)-based methods~\cite{hidasi2016session,xiao2019dynamic} have been proposed. Apart from RNN,  there are also studies that leverage other neural architectures such as  convolutional neural networks (CNN)~\cite{tang2018personalized,yuan2019simple} and Transformer~\cite{kang2018self} to capture sequential patterns. Some attempts formulate the interactive recommendation as  Markov Decision Process (MDP) and solve it via deep Q-learning~\cite{DBLP:conf/kdd/ZhaoZDXTY18,DBLP:conf/www/ZhengZZXY0L18,DBLP:conf/kdd/ZhaoZYLT20,DBLP:conf/sigir/ZhouDC0RTH020} and deep deterministic policy gradient (DDPG)~\cite{zhao2018ac}.  These methods typically focus on  designing novel neural architectures to capture a particular temporal dynamic of user preferences or extract interactive information from user state. As neural architecture design is not the main focus of our work, please refer to these prior works for more details.

\section{Preliminaries}
\textbf{Interactive Recommendation Problem.} We study the interactive (sequential) recommendation problem defined as follows: assume we have a set of users $u \in \mathcal{U},$ a set of items $i \in \mathcal{I}$ and for each user we have access to a sequence of user historical events $\mathcal{E}=\left(i_{1}, i_{2}, \cdots\right)$ ordered by time. Each $i_{k}$ records the item interacted at time $k$. Given the historical interactions, our goal is to recommend to each user a subset of items in order to maximize users' long-term satisfaction.\\
\textbf{Markov Decision Process.}
We translate this sequential recommendation into a MDP 
$\left(\mathcal{S}, \mathcal{A}, \mathbf{P}, R, \rho_{0}, \gamma\right)$ where
$\mathcal{S}$: a continuous state space describing the user states, i.e., $\mathbf{s}_{t}=\left(i_{1}, i_{2}, \cdots, i_{t}\right)$; $\mathcal{A}:$ a discrete action space, containing items available for recommendation;
$\mathbf{P}: \mathcal{S} \times \mathcal{A} \times \mathcal{S} \rightarrow \mathbb{R}$ is the state transition probability;
$R: \mathcal{S} \times \mathcal{A} \rightarrow \mathbb{R}$ is the reward function, where $r(\mathbf{s}_{t}, {a}_{t})$ is the immediate reward obtained by performing action ${a}_{t}$ (the item index) at user state $\mathbf{s}_{t}$. $\rho$ is the initial state distribution; $\gamma$ is the discount factor for future rewards. RL-based RS learns a target policy $\pi_{\bm{\theta}}(\cdot \mid \mathbf{s}_{t})$ which translates the user state $\mathbf{s} \in \mathcal{S}$ into a distribution over all actions $a \in \mathcal{A}$  to maximize  expected cumulative rewards:  
\begin{align}
\max _{\bm{\theta}} \mathbb{E}_{\tau \sim \pi_{\bm{\theta}}}[R(\tau)], \text { where } R(\tau)=\sum_{t=1}^{T} \gamma^{t} r\left(\mathbf{s}_{t}, a_{t}\right). \label{Eq:RL1}
\end{align}
Here the expectation is taken over the trajectories $\tau=\left(\mathbf{s}_{1}, a_{a}, \cdots \mathbf{s}_{T}\right)$ obtained by acting according to the policy: $\mathbf{s}_{0} \sim \rho_{0}(\mathbf{s}_0), a_{t} \sim \pi_{\bm{\theta}}\left(a_t \mid \mathbf{s}_{t}\right)$ and $\mathbf{s}_{t+1} \sim$ $p\left(\mathbf{s}_{t+1} \mid \mathbf{s}_{t}, a_{t}\right)$.\\
\textbf{Offline Learning in Recommendation.} The goal is still to optimize the objective in Eq.~\ref{Eq:RL1}. However, the recommendation policy no longer has the ability to interact with the user in  online environments. Instead, the learning algorithm is provided with a  logged dataset of trajectories, and each trajectory $\tau=\left\{\mathbf{s}_{1}, a_{1}, \cdots, \mathbf{s}_{T}\right\}$ is drawn from a unknown logging (behavior) policy  $\pi_{b}(a_t|\mathbf{s}_t)$. The dataset  can also be represented as  the buffer style $\mathcal{D}=$ $\left\{\left(\mathbf{s}_{t}^{i}, a_{t}^{i}, \mathbf{s}_{t+1}^{i}\right)\right\}_{i=1}^{N}$. For brevity,  we omit  superscript $i$  in what follows.

\section{Analysis on Existing Baselines}
\label{Sec:4}
We first describe the standard offline learning baselines for  interactive recommendation and  analyze their limitations.\\
\textbf{Supervised  Learning.}
The  arguably most direct way to learn interactive RS from logged  feedbacks is to directly apply sequential supervised learning, which  typically relies on the following supervised next-item prediction loss:
\begin{align}
\mathcal{L}_{S}(\bm{\theta})=-\mathbb{E}_{\left(\mathbf{s}_{t}, a_{t}, \mathbf{s}_{t+1}\right)\sim \mathcal{D}}[\log \pi_{\bm{\theta}}\left(a_{t} \mid \boldsymbol{s}_{t}\right)], \label{Eq:S}
\end{align}
where $\pi_{\boldsymbol{\bm{\theta}}}\left(a_{t} \mid \boldsymbol{s}_{t}\right)=\frac{\exp \left(f_{\bm{\theta}}(\mathbf{s}_{t})[a_{t}]\right)}{\sum_{a^{\prime}} \exp \left(f_{\bm{\theta}}(\mathbf{s}_{t})\left[a^{\prime}\right]\right)}$ and $f_{\boldsymbol{\theta}}(\cdot) \in$ $\mathbb{R}^{|\mathcal{A}|}$ denotes a neural network such as RNN~\cite{song2019islf,xiao2019hierarchical}, CNN~\cite{yuan2019simple} or Transformer~\cite{kang2018self}, which is parameterized by $\boldsymbol{\theta}$ and has an output dimension of the action space $|\mathcal{A}|$. $f_{\bm{\theta}}(\mathbf{s}_t)[a_{t}]$ indicates the $a_{t}^{\text {th }}$ index of $f_{\bm{\theta}}(\mathbf{s}_{t})$, i.e., the logit corresponding the  $a_{t}^{\text {th}}$ item.  The supervised learning methods are training stable and easy to implement. However,  they  can not maximize cumulative user rewards.\\
\textbf{Q-learning}. In order to maximize  cumulative user rewards, prior works~\cite{DBLP:conf/www/ZhengZZXY0L18,DBLP:conf/kdd/ZhaoZDXTY18} try to build  RL-based recommendation algorithms by utilizing the deep Q-learning and minimizing the following loss:
\begin{align}
\mathcal{L}_{Q}(\bm{\theta})=\mathbb{E}_{\left(\mathbf{s}_{t}, a_{t}, \mathbf{s}_{t+1}\right)\sim \mathcal{D}}\left( Q_{\bm{\theta}} \left(\mathbf{s}_{t},a_{t} \right)-y \right)^{2},
\end{align}
where the target value $y=r(\mathbf{s}_t,a_t)+\gamma Q_{\bar{\bm{\theta}}}\left(\mathbf{s}_{t+1},\pi \left(\mathbf{s}_{t+1}
\right)\right)$ and  $\pi \left( \mathbf{s}_{t+1} \right)=\arg \max_{a} Q_{\bar{\bm{\theta}}}\left ( \mathbf{s}_{t+1},a\right)$. The $\bar{\bm{\theta}}$ indicates parameters of the target network~\cite{sutton2018reinforcement}.  While the Q-learning  have been widely used in the robot learning literature as an off-policy algorithm~\cite{kumar2019stabilizing}, it is not suitable for offline learning in recommendation task. First, the Q-learning algorithm is affected by extrapolation error~\cite{fujimoto2019off} due to no interaction with environment. Second, in the testing, the robot learning typically  chooses one action according the optimal policy: $\arg \max_{a} Q\left ( \mathbf{s},a\right)$. Instead, in recommendation, we generate  top-k item lists 
by sampling from the softmax function $\frac{Q(\mathbf{s},a)}{\sum_{a'}Q(s,a')}$. This violates the simple machine learning principle: testing and training must match~\cite{vinyals2016matching}. We also find  Q-learning generally does not perform well on logged  feedbacks for  RS in our experiments.

\section{Proposed Offline Learning Framework} 
\subsection{Probabilistic Formulation} 
We start by formally defining the problem of learning  interactive RS in the language of probability. The sequential supervised learning can be expressed as a \textbf{learning} problem of the probabilistic generative model  over observed  $\tau$:
\begin{align}
p(\tau)=\rho(\mathbf{s}_{1})\prod_{t=1}^{T}\pi_{\bm{\theta}}(a_{t}|\mathbf{s}_{t})p(\mathbf{s}_{t+1}|\mathbf{s}_{t}, a_{t}), \label{Eq:SL}
\end{align}
where $\rho(\mathbf{s}_{1})$ and $p(\mathbf{s}_{t+1}\mid\mathbf{s}_{t},\mathbf{a}_{t})$ are the true initial state distribution and dynamics.  
The supervised learning discussed in \S\ref{Sec:4}  can be seen as maximizing the log-likelihood of this generative model respect to $\bm{\theta}$. However, like discussed before, this method can not maximize cumulative rewards. Thus, following previous work~\cite{haarnoja2018latent,levine2018reinforcement},  we introduce a  binary variable $\mathcal{O}$ which are related to rewards by $p(\mathcal{O}_{t}=1|\mathbf{s}_{t},a_t)=\exp (\frac{r(\mathbf{s}_{t},a_{t})}{\alpha})$, where $\alpha$ is a
temperature parameter, to denote whether the action taken at state $\mathbf{s}_{t}$ is optimal. Note that we assume  rewards are nonpositive without loss of generality.  Positive rewards can be scaled  to be no greater than 0.  In the rest of this paper, we use $\mathcal{O}_{t}$ to represent $\mathcal{O}_{t}=1$ for brevity. With this definition, we consider the following generative model with the  trajectory $\tau=\left\{\mathbf{s}_{1}, a_{1}, \cdots, \mathbf{s}_{T}\right\}$ and  variables $\mathcal{O}_{1:T}$:
\begin{align}
p(\tau,\mathcal{O}_{1:T})\propto  \rho(\mathbf{s}_1)\prod_{t=1}^{T}p(\mathbf{s}_{t+1}\mid\mathbf{s}_{t},a_t)p(\mathcal{O}_{t}\mid\mathbf{s}_{t},a_t). \label{Eq:joint}
\end{align}
Note that the action prior is assumed as uniform: $p(a_{t})=\frac{1}{|\mathcal{A}|}$ and is omitted.  The main difference between this formulation and supervised learning (Eq.(\ref{Eq:SL})) is that this formulation considers the optimal trajectory $\tau$ as  latent variable and we focus on the \textbf{inference} problem: inferring the posterior distribution $P(\tau\mid\mathcal{O}_{1:T})$ given optimality of the start  until end of the episode. Based on variational inference~\cite{wainwright2008graphical}, we can derive  the   negative Evidence Lower BOund (ELBO)~\cite{blei2017variational,xiao2019bayesian}  by introducing variational distribution $q_{\bm{\theta}}(\tau)=\rho \left(\mathbf{s}_{1}\right)\prod_{t=1}^{T} p\left(\mathbf{s}_{t+1} \mid \mathbf{s}_{t}, a_{t}\right) \pi_{\bm{\theta}}\left(a_{t} \mid \mathbf{s}_{t}\right)$ (derivations are omitted due to space limitation):
\begin{align}
\mathcal{L}(\bm{\theta})=\mathbb{E}_{\tau \sim q_{\bm{\theta}}(\tau)}\left[\sum_{t=1}^{T}\log q_{\bm{\theta}}(a_t \mid \mathbf{s}_t)- \frac{r\left(\mathbf{s}_{t}, a_{t}\right)}{\alpha}\right]. \label{Eq:ELBO}
\end{align}
Minimizing  the negative ELBO respect to the $\pi_{\bm{\theta}}\left(a_{t} \mid \mathbf{s}_{t}\right)$ is  equal to minimize the Kullback-Leibler (KL) divergence: $\text{KL}(q(\tau)||p(\tau|\mathcal{O}_{1:T}))$.  This objective function can be solved by  soft Q-learning~\cite{haarnoja2017reinforcement} or soft actor-critic algorithms~\cite{haarnoja2018soft}. Although the soft actor-critic is the state-of-the-art off-policy algorithm, it can not be directly applied to RS with discrete actions and it still  suffers from extrapolation error as discussed in the followings. This probabilistic formulation  provides a convenient starting point for our offline learning methods for  recommendation.
\subsection{Inference via Messages Passing} 
\label{sec:inference}
Recall that our goal is  to infer the posterior $p(\tau\mid\mathcal{O}_{1:T})$.
Instead of directly minimizing  Eq. (\ref{Eq:ELBO}), we consider inferring  the posterior through the  message passing algorithm~\cite{heskes2002expectation}. We define the backward messages as:
\begin{align} 
\beta_{t}\left(\mathbf{s}_{t}, a_{t}\right) &:=p\left(\mathcal{O}_{t: T} \mid \mathbf{s}_{t}, a_{t}\right) \label{Eq:m1}\\ 
\beta_{t}\left(\mathbf{s}_{t}\right) &:=p\left(\mathcal{O}_{t: T} \mid \mathbf{s}_{t}\right). \label{Eq:m2}
\end{align}
The recursion for backward messages and optimal policy $\pi_{\bm{\theta}}(a_t\mid\mathbf{s}_{t})$ can be obtained by minimizing Eq. (\ref{Eq:ELBO}) and using  the Markov property of MDP (derivations are omitted due to space limitation):
\begin{align}
 &\beta_{t}\left(\mathbf{s}_{t}, a_{t}\right)  =p\left(\mathcal{O}_{t} \mid \mathbf{s}_{t}, a_{t}\right) \mathbb{E}_{p\left(\mathbf{s}_{t+1}|\cdot \right)}\left[\beta_{t+1}\left(\mathbf{s}_{t+1}\right)\right] \\ &\beta_{t}\left(\mathbf{s}_{t}\right) =\mathbb{E}_{a_{t} \sim p(a_t)}\left[\beta_{t}\left(\mathbf{s}_{t}, a_{t}\right)\right] \\
&\pi_{\bm{\theta}}(a_t\mid\mathbf{s}_{t})= p(a_t\mid\mathbf{s}_{t},\mathcal{O}_{t:T})= \frac{\beta_{t}\left(\mathbf{s}_{t}, a_{t}\right)}{\beta_{t}\left(\mathbf{s}_{t}\right)}\cdot \frac{1}{|\mathcal{A}|}, \label{Eq:Policy}
\end{align}
where $p\left(\mathbf{s}_{t+1}\mid\cdot \right)=p(\mathbf{s}_{t+1}\mid\mathbf{s}_{t},a_t)$ and  $p\left(\mathbf{s}_{1}\mid\cdot \right)=\rho(\mathbf{s}_{1})$. We further use the log form of the backward messages to define the Q value function: $Q\left(\mathbf{s}_{t}, a_{t}\right)=\alpha \log \beta_{t}\left(\mathbf{s}_{t}, a_{t}\right)$. 
In order to conduct inference process, we need to approximate the backward message $Q\left(\mathbf{s}_{t}, {a}_{t}\right)$.   A  direct method is to represent it with parameterized function $Q_{\bm{\phi}}\left(\mathbf{s}_{t}, a_{t}\right)$ with parameter $\bm{\phi}$ and optimize the parameter by minimizing the squared error (derivation is omitted due to space limitation):
\begin{align}
&\mathcal{L}_{Q}(\bm{\phi})=\mathbb{E}_{(\mathbf{s}_{t}, a_{t},\mathbf{s}_{t+1}) \sim q}\big[\frac{1}{2}\big(Q_{\bm{\phi}}\big(\mathbf{s}_{t}, a_{t}\big)-r\left(\mathbf{s}_{t}, a_{t}\right) \label{Eq:Q} \\
&-\left(Q_{\bar{\bm{\phi}}}\left(\mathbf{s}_{t+1}, a_{t+1}\right)- \alpha \log \left(\pi_{\bm{\theta}}\left(a_{t+1} \mid \mathbf{s}_{t+1}\right)\right)\right )\big)^{2}\big] , \nonumber
\end{align}
 The $\bar{\bm{\phi}}$ denotes the parameters of target networks as same as vanilla Q-learning~\cite{hasselt2016deep}.  $q=\pi(\mathbf{s}_{t})\pi_{\bm{\theta}}({a_t\mid\mathbf{s}_{t}})p(\mathbf{s}_{t+1}\mid\mathbf{s}_{t},a_t)$  denotes the variational state-action-state marginal distribution encountered when executing a policy $\pi_{\bm{\theta}}(a_t\mid\mathbf{s}_{t})$. Given the estimated value functions, we can estimate optimal policy according to Eq.~(\ref{Eq:Policy}). Since our policy is discrete, one can directly utilizes the value functions to obtain the non-parametric policy without optimizing $\bm{\theta}$: let $\pi(a_t\mid\mathbf{s}_{t})=\frac{\exp \left (Q_{\bm{\phi}}\left(\mathbf{s}_{t}, a_{t}\right)\right)}{\sum_{a'} \exp \left( Q_{\bm{\phi}}\left(\mathbf{s}_{t}, a'\right)\frac{1}{\alpha}\right)}$. However,  $Q_{\bm{\phi}}\left(\mathbf{s}_{t}, a_{t}\right)$ is typically a very large network in recommendation, e.g. deep ranking model~\cite{covington2016deep}, for alleviating model misspecification, thus it is not computationally efficient  if we use 
 this non-parametric policy. Instead, we consider optimizing a small parametric policy $\pi_{\bm{\theta}}(a_t\mid\mathbf{s}_{t})$ by minimizing the KL divergence  $\text{KL}\big(\pi_{\bm{\theta}}\left(a_t\mid\mathbf{s}_{t}\right)||  \frac{\exp \left (Q_{\bm{\phi}}\left(\mathbf{s}_{t}, a_{t}\right)\right)}{\sum_{a'} \exp \left( Q_{\bm{\phi}}\left(\mathbf{s}_{t}, a'\right)\frac{1}{\alpha}\right)}$, which can be rewritten as  the following loss:
\begin{align}
\mathcal{L}_{\pi}(\bm{\theta})=\mathbb{E}_{a_{t}\sim \pi_{\bm{\theta}}}\left[ \log \pi_{\bm{\theta}}\left(a_{t} \mid \mathbf{s}_{t}\right)- \frac{Q_{\bm{\phi}}\left(\mathbf{s}_{t}, a_{t}\right)}{\alpha} \right], \label{Eq:pi}
\end{align}
where $\pi_{\bm{\theta}}$ is the short for $\pi_{\bm{\theta}}(a_t\mid\mathbf{s}_t)$. This idea is also similar to knowledge distillation~\cite{hinton2015distilling}: we have a large critic (value) network being somewhat "distilled" into a smaller  actor (policy) network. Since $\pi_{\boldsymbol{\bm{\theta}}}\left(a_{t} \mid \boldsymbol{s}_{t}\right)=\frac{\exp \left(f_{\bm{\theta}}(\mathbf{s}_{t})[a_{t}]\right)}{\sum_{a^{\prime}} \exp \left(f_{\bm{\theta}}(\mathbf{s}_{t})\left[a^{\prime}\right]\right)}$ is the   categorical distribution in our case,  a typical
solution for this loss is marginalizing out policy
 over all actions~\cite{christodoulou2019soft}, however, this simple solution is  expensive for a large numbers of items in recommendation. Thus, we address this by utilizing the Gumbel-Softmax~\cite{45822,meng2019semi}, which provides a continuous  differentiable approximation by drawing  samples $y$  from a categorical distribution with class probabilities $f_{\bm{\theta}}(\mathbf{s}_t)$:
\begin{align}
y_{i}=\frac{\exp \left(\left( \log \left( f_{\bm{\theta}}(\mathbf{s}_t)[a_i]\right)+g_{i}\right) / \gamma_g \right)}{\sum_{i=1}^{|\mathcal{A}|} \exp \left(\left ( \log \left (f_{\bm{\theta}}(\mathbf{s}_t)[a_i]\right)+g_{i}\right) / \gamma_g \right)} 
\end{align}
where $\left\{g_{i}\right\}_{i=1}^{|\mathcal{A}|}$ are i.i.d. samples drawn from the Gumbel (0, 1) distribution, $\gamma_g$ is the softmax
temperature and $y_{i}$ is the $i$-th value of sample $\mathbf{y}$. 
Compared to  the Q-learning, our optimal policy $\pi_{\bm{\theta}}({a_{t} \mid \mathbf{s}_{t}})$ has the discrete softmax form not the argmax, which can avoid the mismatch between training and testing. We can iteratively train value function (policy evaluation) Eq.~(\ref{Eq:Q}) and estimate discrete  optimal policy (policy improvement) Eq.~(\ref{Eq:pi}) via stochastic gradient descent. 
\subsection{Stochastic Discrete Actor Critic}
\label{sec:app}
So far, we have proposed a discrete RL algorithm for interactive recommendation. However, we notice that we can not directly apply this algorithm to our offline learning case, since the policy evaluation loss  (Eq.~(\ref{Eq:Q})) can be difficult to optimize due to the dependency between $\pi(\mathbf{s}_{t})$ and $\pi_{\bm{\theta}}({a_t\mid\mathbf{s}_{t}})$
, as well as the need to collect samples from $\pi_{\bm{\theta}}({a_t\mid\mathbf{s}_{t}})$ (on-policy learning).  Thus we consider optimizing
an surrogate approximation $\hat{\mathcal{L}}_{Q}(\bm{\phi})$ of $\mathcal{L}_{Q}(\bm{\phi})$ using the state and action distributions of the unknown logging policy, i.e., $\pi_{b}(\mathbf{s}_{t})$ and $\pi_{b}(a_t\mid \mathbf{s}_{t})$. In other words, we can use the logged  feedbacks to conduct the policy evaluation step: 
\begin{align}
&\hat{\mathcal{L}}_{Q}(\bm{\phi})=\mathbb{E}_{(\mathbf{s}_{t}, {a}_{t}, \mathbf{s}_{t+1}) \sim \mathcal{D}}\big[\frac{1}{2}\big(Q_{\bm{\phi}}\big(\mathbf{s}_{t}, a_{t}\big)-r(\mathbf{s}_{t},a_t) \label{Eq:SQ} \\
&-\left(Q_{\bar{\bm{\phi}}}\left(\mathbf{s}_{t+1}, a_{t+1}\right)- \alpha \log \left(\pi_{\bm{\theta}}\left(a_{t+1} \mid \mathbf{s}_{t+1}\right)\right)\right)\big)^{2}\big] . \nonumber
\end{align}
$\hat{\mathcal{L}}_{Q}(\bm{\phi})$ matches $\mathcal{L}_{Q}(\bm{\phi})$ to first order approximation~\cite{kakade2002approximately}, and provides a reasonable estimate of $\mathcal{L}_{Q}(\bm{\phi})$ if $\pi_{b}(a_t\mid\mathbf{s}_{t})$ and $\pi_{\bm{\theta}}({a_t\mid\mathbf{s}_{t}})$ are similar. However, we find it performs dramatically worse than supervised learning methods on logged implicit feedbacks  in our recommendation task  due to the extrapolation error~\cite{fujimoto2019off}.  As discussed in the prior work~\cite{fujimoto2019off}, extrapolation error can be attributed to a mismatch in the distribution of logged feedbacks induced by the policy $\pi_{\bm{\theta}}(a_t\mid\mathbf{s}_{t})$
and the distribution $\pi_{b}(a_t\mid\mathbf{s}_t)$ of data contained in the batch in the policy iteration process. It is therefore crucial to control divergence of the optimizing  policy $\pi_{\bm{\theta}}(a_t\mid \mathbf{s}_{t})$ and the unknown  logging policy $\pi_{b}(a_t\mid \mathbf{s}_t)$.  To this effect, we explore five approaches for our discrete stochastic actor-critic algorithm in the followings.
 
\subsection{Support Constraints}
\label{Sec:sc}
Previous work ~\cite{fujimoto2019off} extends Q-learning and tries to  disallow any action that has zero support under the logging policy,  which can be formally defined as the $\delta$-behavior constrained policy:
 \begin{align}
 \mathbb{1}\left[a_{t}=\arg \max _{a}\left\{\hat{Q}\left(\mathbf{s}_{t},a\right): \pi_{b}\left(a \mid \mathbf{s}_{t} \right) \geq \delta \right\}\right],
 \end{align}
 where $0\leq \delta \leq 1$. This  constrains the deterministic argmax greedy policy in vanilla Q-learning to actions that the logging policy chooses with probability at least $\delta$. However, this constraint is not suitable for  our proposed stochastic softmax policy. Thus, we  combine the policy improvement loss $\mathcal{L}_{\pi}(\bm{\theta})$ (Eq.~(\ref{Eq:pi})) and the support constraint  as a joint loss: $\mathcal{L}_{\pi}^{\text{sc}}(\bm{\theta})=\mathcal{L}_{\pi}(\bm{\theta})+\beta \mathcal{R}_{\text{sc}}(\bm{\theta})$, where $\beta$ is the coefficient  and the proposed support constraint term  $\mathcal{R}_{sc}(\bm{\theta})$ is:
 \begin{align}
 \mathcal{R}_{\text{sc}}(\bm{\theta})=-\sum_{a \in \mathcal{A}, \pi_{b}(a \mid \mathbf{s}_{t}) \leq \delta }  \pi_{\bm{\theta}}(a \mid \mathbf{s}_{t}). \label{Eq:SC}
 \end{align}
This  method is very simple and conceptually straightforward but requires estimating the unknown logging policy when $\delta>0$. In addition, it is just one-step constraint of policy at time step $t$, meaning that it can not avoid actions that may lead to higher deviation at future time steps. 
 
\subsection{Supervised Regularization}
\label{sec:sr}
In order to avoid estimating the logging policy, we propose  a auxiliary supervised regularization to reduce the extrapolation error by minimizing the forward KL divergence, i.e., $\text{KL}\big(\pi_{b}(a_{t}\mid\mathbf{s}_{t})||\pi_{\bm{\theta}}({a_t\mid\mathbf{s}_{t}})\big)$ between the unknown logging policy $\pi_{b}(a_{t}\mid\mathbf{s}_{t})$ and recommendation policy $\pi_{\bm{\theta}}({a_t\mid\mathbf{s}_{t}})$:
\begin{align}
\mathcal{R}_{\text{sr}}(\bm{\theta})=-\mathbb{E}_{(a_{t} ,\mathbf{s}_t)\sim \pi_{b}(a_{t},\mathbf{s}_{t})}\left[\log \pi_{\bm{\theta}}\left(a_{t} \mid \boldsymbol{s}_{t}\right) \right].\label{Eq:Rsr}
\end{align}
This method does no require a pre-estimated logging policy since we can directly use the logged data drawn from $\pi_{b}(a_{t}\mid\mathbf{s}_{t})$. Similar to support constraints, we need combine the policy improvement loss  and the supervised loss: $\mathcal{L}_{\pi}^{\text{sr}}(\bm{\theta})=\mathcal{L}_{\pi}(\bm{\theta})+\beta \mathcal{R}_{\text{sr}}(\bm{\theta})$. However, the two terms are on an arbitrary relative scale, which can make it difficult to choose $\beta$. To adaptively update $\beta$, we  replace the soft supervised  regularization with a hard constraint with parameter $\epsilon$. The optimization loss for the policy improvement becomes:
\begin{align}
\mathcal{L}&_{\pi}^{\text{sr}}(\bm{\theta})=\min_{\bm{\theta}}\mathbb{E}_{a_{t}\sim \pi_{\bm{\theta}}}\left[\log \pi_{\bm{\theta}}\left(a_{t} \mid \mathbf{s}_{t}\right)- \frac{Q_{\bm{\phi}}\left(\mathbf{s}_{t}, a_{t}\right)}{\alpha} \right], \nonumber \\
&\text {s. t.} \quad \mathbb{E}_{\mathbf{s}_{t} \sim \pi_{b}(\mathbf{s}_{t}) }\left[\operatorname{KL}\left[\pi_{b} (a_{t} \mid \mathbf{s}_{t}) \| \pi_{\bm{\theta}}(a_t\mid\mathbf{s}_{t})\right]\right] \leq \epsilon.  \label{Eq:ac}
\end{align}
To optimize this, we  introduce a Lagrange multiplier $\beta$  via dual gradient descent and Gumbel-Softmax (derivations are omitted due to space limitation):
\begin{align}
&\mathcal{L}_{\pi}^{\text{sr}}(\bm{\theta},\beta)=\min_{\bm{\theta}}\max_{\beta \geq  0}\mathbb{E}_{a_{t}\sim \pi_{\bm{\theta}}}\left[\alpha \log \pi_{\bm{\theta}}\big(a_{t} \mid \mathbf{s}_{t}\right)-   \\
&Q_{\bm{\phi}}\big(\mathbf{s}_{t}, a_{t}\big) \big] -\beta \left(\mathbb{E}_{\mathbf{s}_{t} \sim \pi_{b} }[\operatorname{KL}[\pi_{b} (a_{t} \mid \mathbf{s}_{t}) \| \pi_{\bm{\theta}}(a_t \mid \mathbf{s}_{t})]] - \epsilon \right). \nonumber
\end{align}
As discussed in \S\ref{sec:app} and demonstrated in our experiments, enforcing a specific constraint between learning policy and logging policy is critical for good performance.  
\subsection{Policy Constraints}
\label{sec:pc}
Since the KL divergence is asymmetric, we can also consider constraining the policy via the  reverse KL divergence $\text{KL}\big(\pi_{\bm{\theta}}({a}_{t}\mid\mathbf{s}_{t}) \|\pi_{b}(a_{t}\mid\mathbf{s}_{t})\big)$ during our  policy improvement process. To see this, we
add the reverse KL constraint to Eq.~(\ref{Eq:pi})  and remove the negative entropy term $\pi_{\bm{\theta}}({a}_{t}\mid \mathbf{s}_{t})\log \pi_{\bm{\theta}}({a}_{t}\mid \mathbf{s}_{t})$ in Eq.~(\ref{Eq:pi}) since there is already a entropy term in the reverse KL divergence. Then, the objective for the policy improvement with policy constraint is:
\begin{align}
\mathcal{L}_{\pi}^{\text{pc}}(\bm{\theta})&=\min_{\bm{\theta}}\mathbb{E}_{a_{t}\sim\pi_{\bm{\theta}}}\left[-Q_{\bm{\phi}}\left(\mathbf{s}_{t}, a_{t}\right) \right], \label{Eq:PC}  \\
\text {s. t.} &\quad \mathbb{E}_{\mathbf{s}_{t} \sim \pi_{b}(\mathbf{s}_{t}) }\left[\operatorname{KL}\left[\pi_{\bm{\theta}} (a_{t} \mid \mathbf{s}_{t}) \| \pi_{b}(a_t\mid\mathbf{s}_{t})\right]\right] \leq \epsilon. \nonumber 
\end{align}
Similar to Supervised Regularization, optimizing this constrained   problem via introducing Lagrange multiplier $\beta$ results the final policy improvement objective:
\begin{align}
\mathcal{L}_{\pi}^{\text{pc}}(\bm{\theta})&=\min_{\bm{\theta}} \mathbb{E}_{(a_t, \mathbf{s}_t) \sim \pi_{b}(a_t, \mathbf{s}_{t})}\left[w_t \log \pi_{\theta}(a_{t} \mid \mathbf{s}_{t}) \right],
\end{align}
where $w_t=- \exp \left(\frac{Q_{\bm{\phi}}(\mathbf{s}_t,a_t)}{\beta}\right)$. This objective simplifies the policy improvement problem to a weighted maximum likelihood problem, and does not require an estimated logging policy and Gumbel-Softmax approximation.\subsection{Dual Constraints}
\label{sec:dc}
Although the methods proposed in previous subsections can effectively   control the divergence of  optimizing policy and  data logging policy,  both of them only consider one-step regularization, and thus can not avoid actions that may lead to higher deviation at future time steps. To address this problem, instead of explicitly  adding constraint for policy improvement at time $t$, we consider the logging policy as a prior and incorporating it directly 
 into our original  probabilistic model (Eq.~\ref{Eq:joint}). Specifically, we first estimate the logging policy $\hat{\pi}_{b}(a_{t}\mid\mathbf{s}_{t})$ via supervised learning using the logged feedbacks as the proxy of unknown logging ${\pi}_{b}(a_{t}\mid\mathbf{s}_{t})$, and then incorporate it as the action prior into Eq.~(\ref{Eq:joint}), which yields the following joint distribution:
\begin{align}
 p=\rho(\mathbf{s}_1)\prod_{t=1}^{T}p(\mathbf{s}_{t+1}\mid \mathbf{s}_{t},\mathbf{a}_t)p(\mathcal{O}_{t} \mid \mathbf{s}_{t},\mathbf{a}_t)\hat{\pi}_{b}(a_{t}|\mathbf{s}_{t}),
\end{align}
where $p$ denotes the joint distribution $p(\tau,\mathcal{O}_{1:T})$. Given this joint distribution, similar to \S\ref{sec:inference}, we can infer the posterior $p(\tau\mid\mathcal{O}_{1:T})$ by introducing the variational distribution $q_{\bm{\theta}}(\tau)=\rho \left(\mathbf{s}_{1}\right)\prod_{t=1}^{T} p\left(\mathbf{s}_{t+1} \mid \mathbf{s}_{t}, a_{t}\right) \pi_{\bm{\theta}}\left(a_{t} \mid \mathbf{s}_{t}\right)$.  We also consider  inferring the posterior through the message passing algorithm as same as we do in    \S\ref{sec:inference}. However, since the action prior is no longer uniform, we have two choices of definitions of the backward messages. The first is defined as same as Eq. (\ref{Eq:m1}) and Eq. (\ref{Eq:m2}), i.e, $\beta_{t}\left(\mathbf{s}_{t}, a_{t}\right) :=p\left(\mathcal{O}_{t: T} \mid \mathbf{s}_{t}, a_{t}\right)$ and $\beta_{t}\left(\mathbf{s}_{t}\right) :=p\left(\mathcal{O}_{t: T}\mid\mathbf{s}_{t}\right).$ Based on this definition and through minimizing $\text{KL}\left(q_{\bm{\theta}}\left(\tau\right)\| p\left(\tau\mid\mathcal{O}_{1:T}\right)\right)$, we can obtain following objectives for policy evaluation  and policy improvement:
\begin{align}
&\mathcal{L}_{Q}^{\text{dc}}(\bm{\phi})=\mathbb{E}_{(\mathbf{s}_{t}, a_{t}, \mathbf{s}_{t+1}) \sim \mathcal{D}}\big[\frac{1}{2}\big(Q_{\bm{\phi}}\big(\mathbf{s}_{t}, a_{t}\big)-r(\mathbf{s}_{t},a_t) \label{Eq:DC}  \\
&-\big(Q_{\bar{\bm{\phi}}}\left(\mathbf{s}_{t+1}, a_{t+1}\right)+\alpha \log \hat{\pi}_{b}(a_{t+1}\mid\mathbf{s}_{t+1}) \\
&-\alpha\log \left( \pi_{\bm{\theta}}\left(a_{t+1} \mid \mathbf{s}_{t+1}\right)\right)\big )\big)^{2}\big] , \\
&\mathcal{L}_{\pi}^{\text{dc}}(\bm{\theta})=\mathbb{E}_{a_{t}\sim \pi_{\bm{\theta}}}\big[\log  \frac{\pi_{\bm{\theta}}\left(a_{t} \mid \mathbf{s}_{t}\right)}{\hat{\pi}_{b}(a_t\mid\mathbf{s}_t)}- \frac{Q_{\bm{\phi}}\left(\mathbf{s}_{t}, a_{t} \right)}{\alpha} \big]. \label{Eq:DCPi}
\end{align}
Compared to the supervised regularization in \S\ref{sec:sr} and policy constraint in \S\ref{sec:pc}, this method  not only adds prior constraint on the policy improvement step, but also adds it to the target Q value, which  can avoid actions that are far from logging policy at future time steps.  \\
\subsection{Reward Extrapolation}
Following the dual constraints in \S\ref{sec:dc}, if we consider  the second definition of the backward messages: $\beta_{t}\left(\mathbf{s}_{t}, a_{t}\right) :=p\left(\mathcal{O}_{t: T} \mid \mathbf{s}_{t}, a_{t}\right)\hat{\pi}_b({a_t\mid \mathbf{s}_t})$ and $\beta_{t}\left(\mathbf{s}_{t}\right) :=p\left(\mathcal{O}_{t: T} \mid \mathbf{s}_{t}\right)$, we can obtain following objectives for policy iteration:
\begin{align}
&\mathcal{L}_{Q}^{\text{re}}(\bm{\phi})=\mathbb{E}_{(\mathbf{s}_{t}, a_{t}, \mathbf{s}_{t+1}) \sim \mathcal{D}}\big[\frac{1}{2}\big(Q_{\bm{\phi}}\big(\mathbf{s}_{t}, a_{t}\big)-\hat{r}(\mathbf{s}_t, a_t) \\
&-\big(Q_{\bar{\bm{\phi}}}\left(\mathbf{s}_{t+1}, a_{t+1}\right)-\alpha \log \left(\pi_{\bm{\theta}}\left(a_{t+1} \mid \mathbf{s}_{t+1}\right)\right)\big )\big)^{2}\big] , \\
&\mathcal{L}_{\pi}^{\text{re}}(\bm{\theta})=\mathbb{E}_{a_{t}\sim \pi_{\bm{\theta}}}\big[\log \pi_{\bm{\theta}}\left(a_{t} \mid \mathbf{s}_{t}\right)- \frac{Q_{\bm{\phi}}\left(\mathbf{s}_{t}, a_{t} \right)}{\alpha} \big],
\end{align}
where $\hat{r}(\mathbf{s}_t, a_t)=r(\mathbf{s}_t, a_t)+\alpha \log \hat{\pi}_{b}(a_{t} \mid \mathbf{s}_{t})$.  We can observe that this method  extrapolates the task specific rewards with the output of the estimated logging policy $\hat{\pi}_{b}(a_t\mid\mathbf{s}_{t})$. This modified objective forces the model to learn that the most valuable actions, but still have high probability in the original logged feedbacks. Note that while  inference objectives for dual constraints and reward extrapolation are same, the particular choice of how the backward messages are defined  can make a significant difference in practice due to the difference between actor and critic in network architectures. 
\section{Experiments}
In this section, we empirically analyze and compare the effectiveness  of the proposed approaches. We conduct  experiments on two public real-world datasets and investigate the following research questions: \textbf{(RQ1)} How do the proposed methods perform compared with existing   methods for interactive recommendation? 
\textbf{(RQ2)} Are the proposed learning methods robust to different types of neural architectures and sparse logged feedbacks?
\textbf{(RQ3)} Can the adaptive update step improve performance of the supervised regularization? \textbf{(RQ4)} How sensitive is the performance of the proposed learning  methods with respect to the trade-off parameters?
\begin{table}
\center
\setlength{\tabcolsep}{1.5mm}{
\resizebox{0.45\textwidth}{!}{
\begin{tabular}{@{}c cccc c cccc c  @{}}
\toprule[1.0pt]
  & \multicolumn{4}{c}{RecSys}  & & \multicolumn{4}{c}{Kaggle}   \\

  \cline{2-5}    \cline{7-10} 
  \cline{2-5}    \cline{7-10} 
     &  H$@$5  & N$@$5  & H$@$10   & N$@$10  & &  H$@$5  & N$@$5  & H$@$10   & N$@$10  &  \\
     \toprule[1.0pt]
     
   SL    &  .2876  & .1982  & .3793   & .2279  & &  .2233  & .1735  & .2673   & .1878  &  \\
       \toprule[1.0pt]
   DQN   &  .2134  & .1215  & .3125   & .1673  & &  .1471  & .0953  & .1965   & .1176  &  \\

   PG   &  .2151  & .1279  & .3218   & .1792  & &  .1585  & .1041  & .2083   & .1212  &  \\
       \toprule[1.0pt]
   SL+DQN    &  .2991  & .2012  & .3951   & .2348  & &  .2487  & .1939  & .2967   & .2094  &  \\

   SL+PG    &  .3012  & .2106  & .4013   & .2382  & &  .2504  & .1972  & .3036   & .2118  &  \\
   	\toprule[1.0pt]
   SDAC    &  .2341  & .1332  & .3316   & .1872  & &  .1669  & .1162  & .2173   & .1358  &  \\
    \toprule[1.0pt]
   SC    &  .2987  & .1991  & .3905   & .2356  & &  .2352  & .1885  & .2854   & .1962  &  \\
     
   SR    &    \underline{.3197}   & \underline{.2234}   & \underline{.4184}  &  \underline{.2515}  & &   \underline{.2586}  &   \underline{.2087}   & \underline{.3153}   & \underline{.2259}  &  \\

   PC    &  .3081  & .1986  & .3903   & .2319  & &  .2354  & .1913  & .2941   & .1958  &  \\

   DC    &  $\textbf{.3272}^{\ast}$  & $\textbf{.2306}^{\ast}$  & $\textbf{.4217}^{\ast}$   & $\textbf{.2593}^{\ast}$  & & $ \textbf{.2659}^{\ast}$  & $\textbf{.2181}^{\ast}$  & $\textbf{.3204}^{\ast}$   & $\textbf{.2351}^{\ast}$  &  \\

   RE   &  .3128  & .2195   & .4071   & .2416  & &  .2528  & .2043  & .3085   & .2192  &  \\
   
  \bottomrule[1.0pt]
\end{tabular}}}
\caption{Performance comparison of difference learning algorithms utilizing  RNN as the backbone. The best and the second best performance  are marked with boldfaces and underlined, respectively. $\ast$ indicates the method outperforms  others at a significance level of $p\leq 0.01$ by paired t-tests.}
\label{Table:overall}
\end{table}
\subsection{Experimental Setup}
We use  following two public real-world datasets: \\
\textbf{RecSys}~\footnote{https://recsys.acm.org/recsys15/challenge/}: This dataset is a public dataset  released by RecSys
Challenge 2015 and contains sequences of user purchases and clicks. After preprocessing,  it contains 200,000 session sequence and 1,110,965 interactions over 26,702 items. \\
\textbf{Kaggle}~\footnote{https://www.kaggle.com/retailrocket/ecommerce-dataset}: This dataset  comes from a real-world e-commerce website. After preprocessing, it contains 195,523 sequence and 1,176,680 interactions over 70,852 items.\\
\textbf{Evaluation Metrics}. For offline evaluation, we employ top-k Hit Ratio (HR$@$k) and  Normalized Discounted Cumulative Gain (NDCG$@$k) to evaluate the  performance, which are widely used in related works~\cite{ChenWang21,ChenDTTD2021,ChenGW19,xiao2019neural}. We report results on HR (H)$@$\{5, 10\} and NDCG (N)$@$\{5, 10\}. We adopt cross-validation to evaluate
the performance of the proposed methods. we randomly sample 80\% sequences as the training set, 10\% as validation and the rest as test set. We use all items as candidates and rank them for evaluation. Each experiment is repeated 5 times, and the average performance is reported. \\
\textbf{Baselines}. Since this paper focuses on proposing  learning algorithms. We consider  following learning algorithms  as our  baselines:
Supervised Learning (SL), Deep Q-Learning (DQN),  Off-Policy Gradient (PG)~\cite{DBLP:conf/wsdm/ChenBCJBC19},  (SL+DQN)~\cite{DBLP:conf/sigir/XinKAJ20}, (SL+PG)~\cite{DBLP:conf/kdd/GongZDLGSOZ19}. 
We compare  these baselines with our proposed stochastic discrete actor critic (SDAC), support constraints (SC), supervised regularization (SR), policy constraints (PC), dual constraints (DC) and reward extrapolation (RE).
All  learning methods are based on same backbone i.e., recurrent neural networks (RNN) introduced by work~\cite{DBLP:conf/wsdm/ChenBCJBC19}, in order to avoid the choice of backbone to be a confounding factor. Hyperparameters  are tuned on validation set.
\begin{table}
\center
\setlength{\tabcolsep}{1.5mm}{
\resizebox{0.45\textwidth}{!}{
\begin{tabular}{@{}c cccc c cccc c  @{}}
\toprule[1.0pt]
  & \multicolumn{4}{c}{RecSys}  & & \multicolumn{4}{c}{Kaggle}   \\

  \cline{2-5}    \cline{7-10} 
  \cline{2-5}    \cline{7-10} 
     &  H$@$5  & N$@$5  & H$@$10   & N$@$10  & &  H$@$5  & N$@$5  & H$@$10   & N$@$10  &  \\
     \toprule[1.0pt]
   
   SL    &  .2728  & .1896  & .3593   & .2177  & &  .1966  & .1566  & .2302   & .1675  &  \\
       \toprule[1.0pt]
   DQN   &  .1946  & .1075  & .3004   & .1562  & &  .1132  & .0621  & .1323   & .0958  &  \\
    PG   &  .2031  & .1191  & .3079   & .1616  & &  .1212  & .0696  & .1428  & .1024  &  \\
       \toprule[1.0pt]
   SL+DQN    &  .2742  & .1909  & .3613   & .2192  & &  .2089  & .1611  & .2454   & .1778  &  \\
    SL+PG   &  .2776  & .1977  & .3678   & .2215  & &  .2107  & .1747  & .2504  & .1804  &  \\
       \toprule[1.0pt]
   SDAC    &  .2138  & .1207  & .3109   & .1643  & &  .1342  & .0927  & .1566   & .1212  &  \\
    \toprule[1.0pt]
   SC    &  .2665  & .1913  & .3702   & .2202  & &  .2007  & .1576  & .2369   & .1772  &  \\
     
   SR    &  .2841  & .2076  & .3741  &  .2311 & &  .2241  & .1798  & .2538   & \underline{.1895}  &  \\

   PC    &  .2782  & .1918  & .3692   & .2214  & &  .2132 & .1752  & .2498   & .1831  &  \\

   DC    &  $\textbf{.2951}^{\ast}$  & $\textbf{.2136}^{\ast}$  & $\textbf{.3853}^{\ast}$   & $\textbf{.2413}^{\ast}$  & & $ \textbf{.2341}^{\ast}$  & $\textbf{.1853}^{\ast}$  & $\textbf{.2745}^{\ast}$   & $\textbf{.1920}^{\ast}$  &  \\

   RE    &  \underline{.2866}  & \underline{.2101}  & \underline{.3847}   & \underline{.2371}  & &  \underline{.2289}  & \underline{.1801}  & \underline{.2673}   & .1866  &  \\
  \bottomrule[1.0pt]
\end{tabular}}}
\caption{Performance comparison of difference learning algorithms utilizing  CNN as the backbone. The best and the second best performance  are marked with boldfaces and underlined, respectively. }
\label{Table:robust}
\end{table}
\subsection{Experimental Results}
\begin{figure}[!t]
\centering
  \subfigure{
    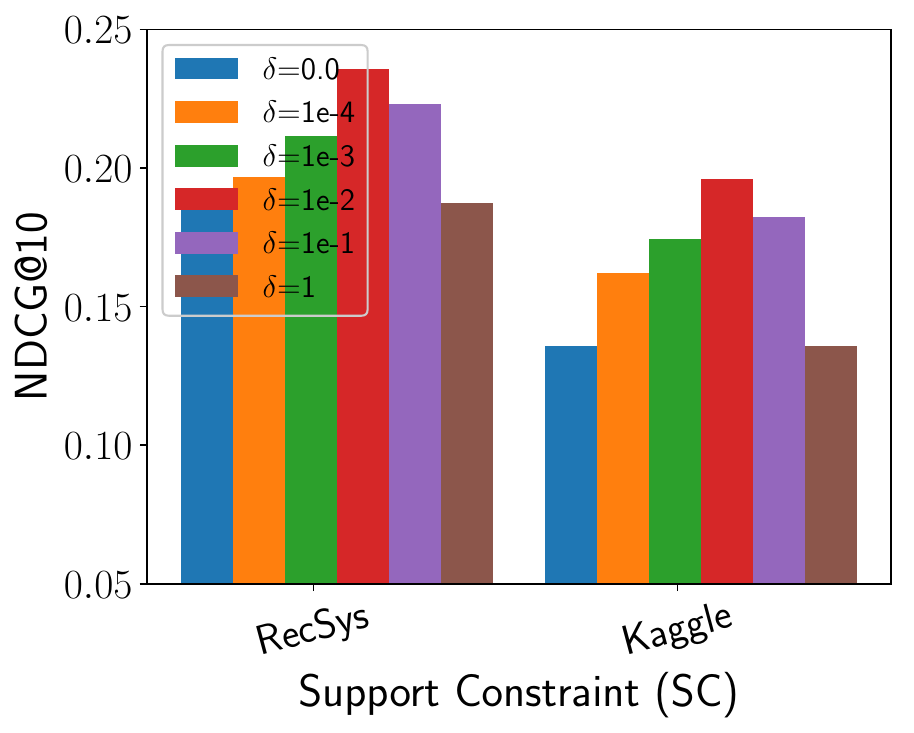
\label{subfig:SCNDCG.pdf}
    \includegraphics[width=0.228\textwidth]{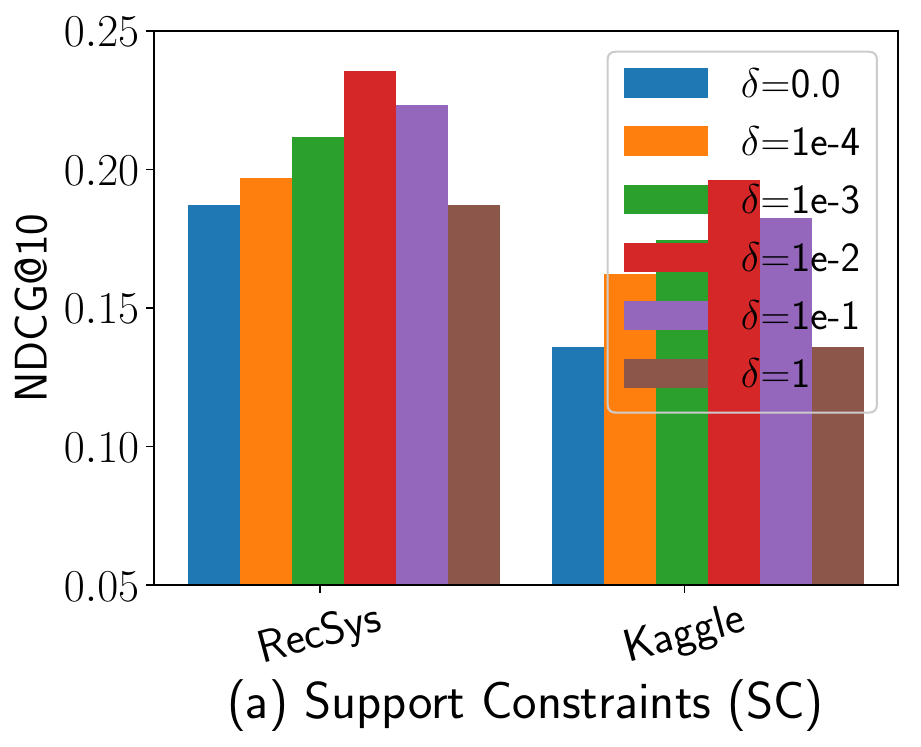}}
  \subfigure{
    \label{subfig:PCNDCG}
    \includegraphics[width=0.228\textwidth]{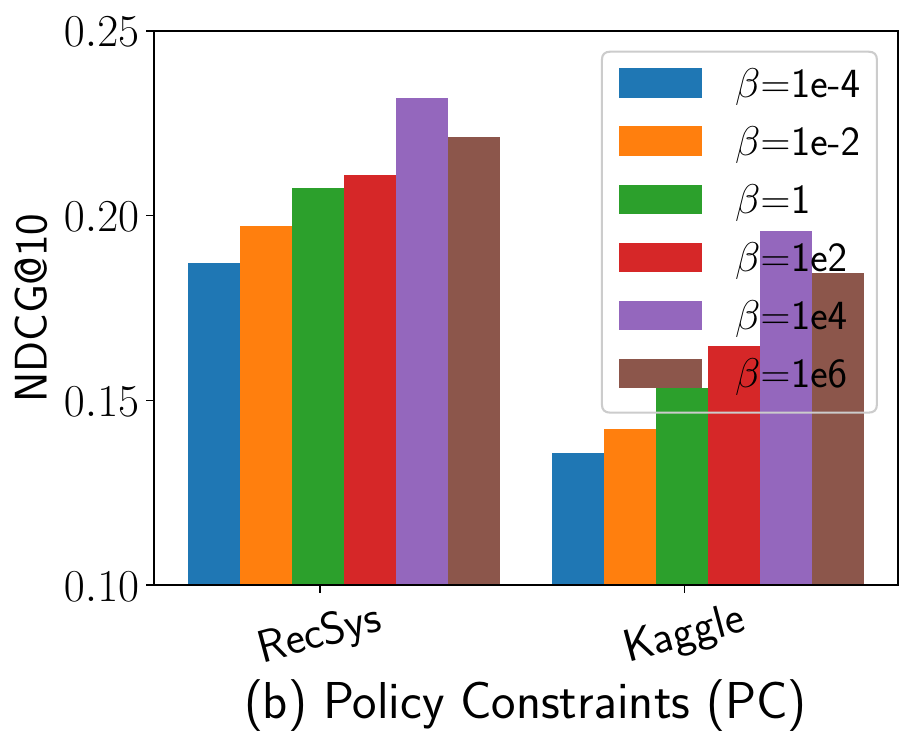}}
      \subfigure{
    \label{subfig:DCNDCG}
    \includegraphics[width=0.228\textwidth]{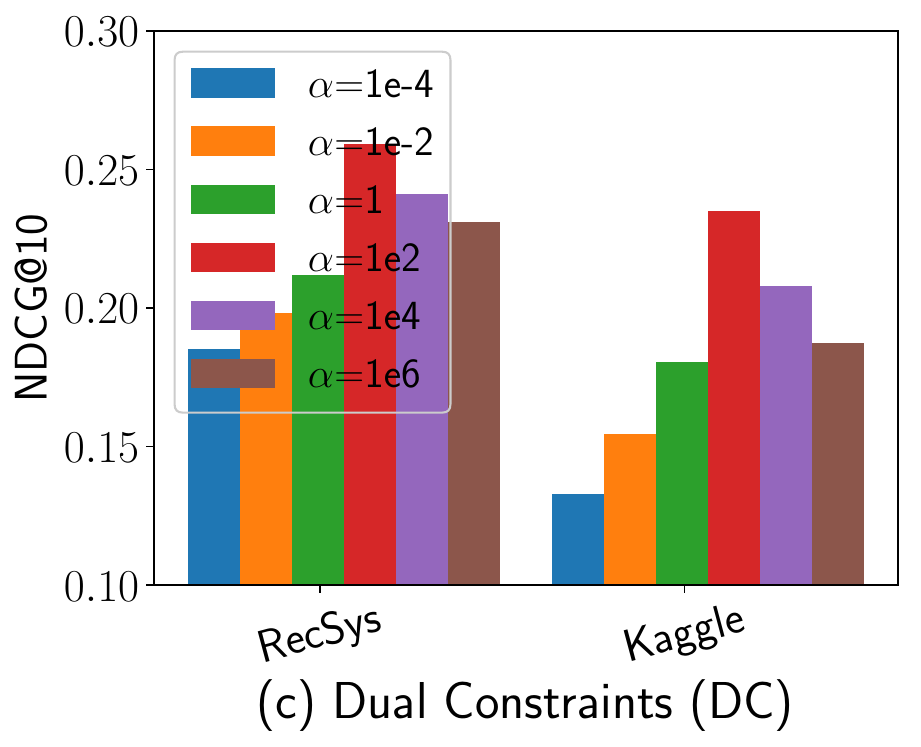}}
  \subfigure{
    \label{subfig:RANDCG}
    \includegraphics[width=0.228\textwidth]{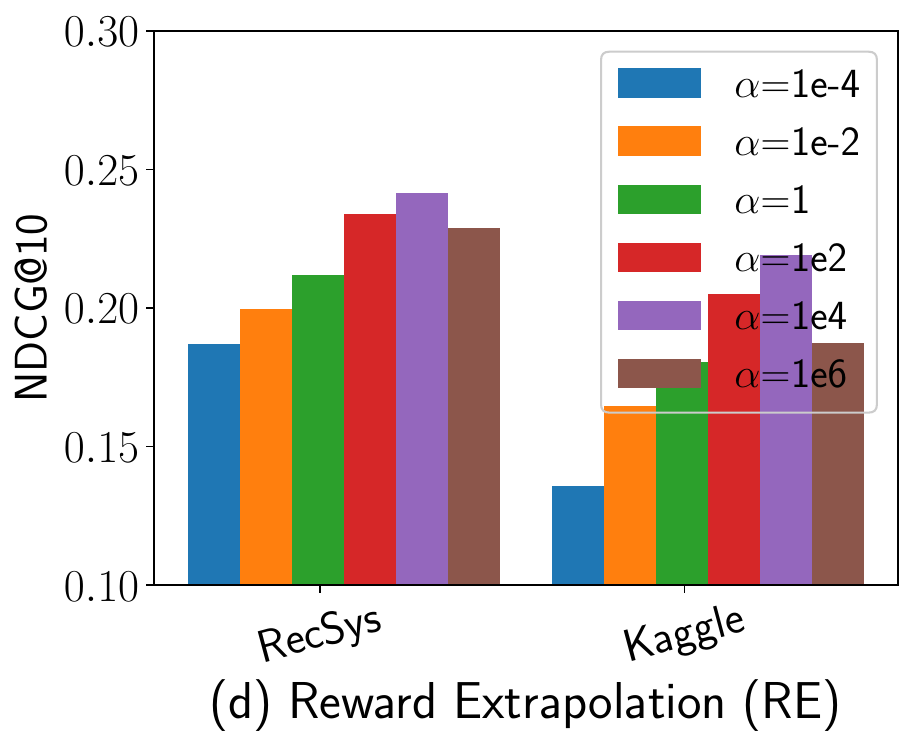}}
    \caption{Performance comparison of NDCG when varying the trade-off parameters in four methods on two datasets.}\label{fig:sen}
\end{figure}

\textbf{Overall Performance (RQ1)}.
Table~\ref{Table:overall} shows the  performance of our proposed methods and the baselines. From this table,  we have the following observations: (a) The off-policy RL algorithms (DQN and PG) perform dramatically worse than SL, demonstrating that off-policy RL algorithms can not effectively learn a optimal policy without online interactions due to the extrapolation error. (b) 
Our SDAC  outperform DQN, though both of them are off-policy RL algorithms. One possible reason is that our SDAC learns  a stochastic discrete policy, which make it more suitable for recommendation task with discrete items compared to DQN. (c) The proposed methods with policy constraints or regularization, e.g., SR, RE and DC significantly outperform the proposed SDAC, which demonstrates that  minimizing the mismatch between recommendation policy and logging policy is important when training an off-policy RL algorithm in the offline setting. (d) Our offline learning methods RE and DC outperform the SL, indicting that exploiting both supervision and task reward, and maximizing  cumulative rewards  does help improve the recommendation performance. \\
\textbf{Different backbones and sparse feedbacks (RQ2)}.
In the \textbf{RQ1}, we implement all methods based on the RNN backbone. To further evaluate the effectiveness of the proposed offline learning methods, we  consider the case where the learning methods is implemented using other neural network architectures. We consider other two state-of-the-art neural architectures in recommendation as backbones, i.e., temporal CNN (Caser)~\cite{tang2018personalized} and Transformer (SASRec)~\cite{kang2018self}. We only show the results on CNN  due to space limitations, and the results of Transformer are similar to CNN and are thus omitted.  As shown in Table~\ref{Table:robust}, our offline learning methods perform better than all the compared methods in most cases, which  once again proves the effectiveness of our methods and also shows that our offline learning methods is robust to different backbones.
\begin{figure}[h]
\centering
  \subfigure{
    \label{subfig:betaHR}
    \includegraphics[width=0.228\textwidth]{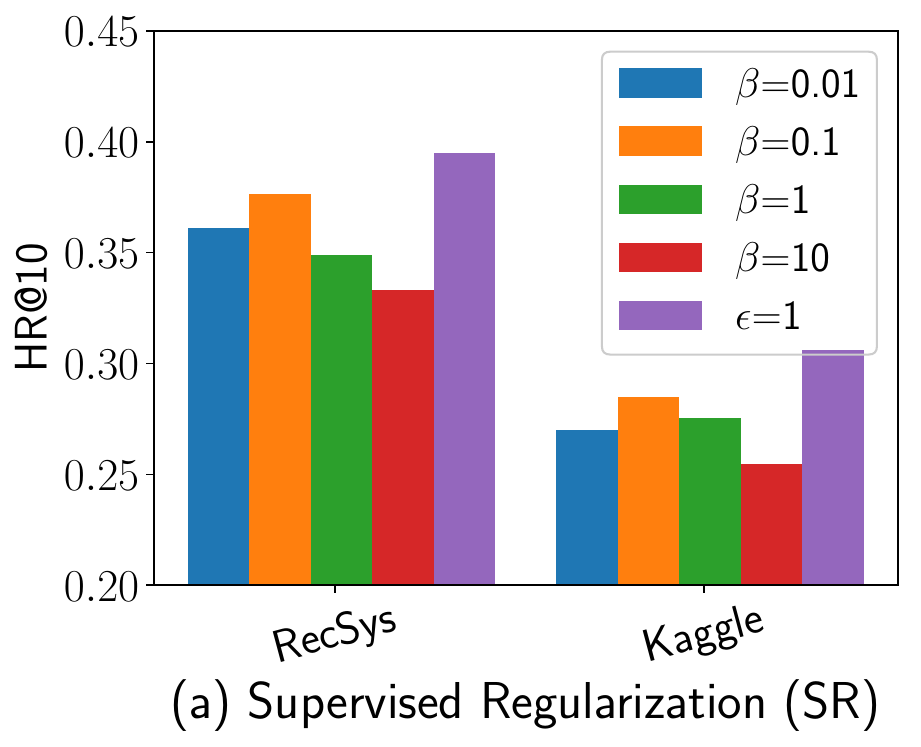}}
  \subfigure{
    \label{subfig:betaNDCG}
    \includegraphics[width=0.228\textwidth]{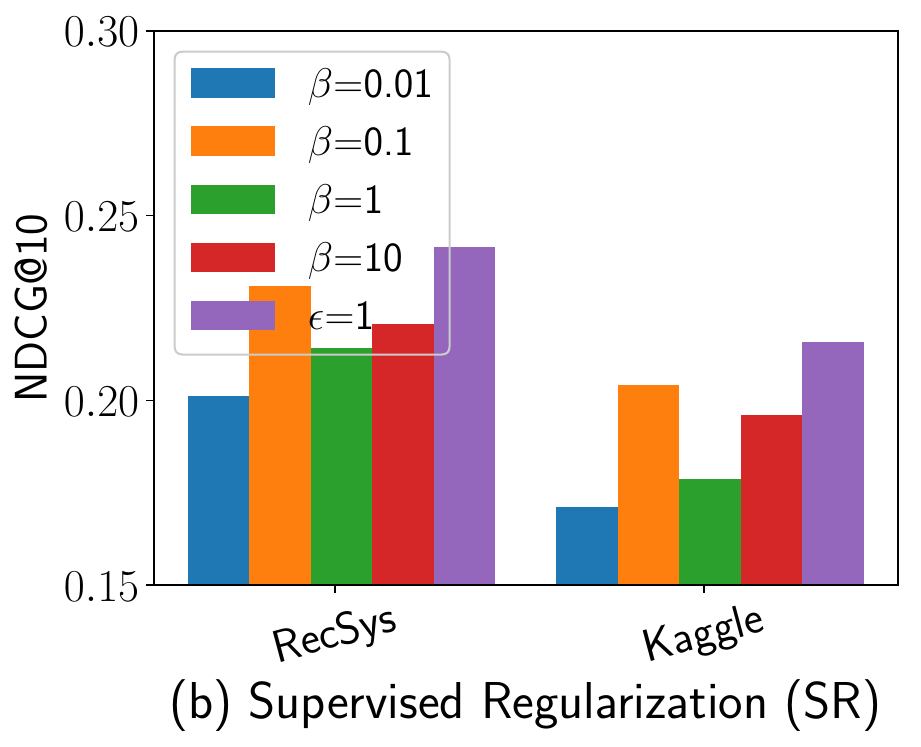}}
\caption{Comparison of supervised regularization   on two datasets with different $\beta$ utilizing  RNN as backbone. }\label{fig:beta}
\end{figure}
To evaluate the effectiveness of proposed offline learning methods on more sparse logged feedbacks, we consider the purchase,  a more sparse feedback compared to click. RecSys dataset contains 43,946 purchases of users. For Kaggle dataset, we consider the behavior of adding to cart as the purchase feedback, resulting 57,269 purchase feedbacks. Table~\ref{Table:sparse} shows the purchase performance comparison of different learning algorithms. 
According to the table: (1) Our offline learning methods such as DC, SR and RE still dominate other  baselines, which confirms that our methods can also perform well on  sparse logged feedbacks.  (2) Different from the Table~\ref{Table:overall}, RE becomes a competitive method and outperforms  SR, which demonstrates that control the divergence of recommendation policy and  logging policy at future time steps is helpful on sparse feedbacks.\\
\textbf{Adaptive Regularization (RQ3)} To evaluate the effects of the adaptive $\beta$ updates in the SR method (see \S\ref{sec:sr}), we compare policies trained with different fixed values of $\beta$ and  policies where $\beta$ is updated adaptively to enforce a desired distribution constraint $ \epsilon= 1$.  Figure~\ref{fig:beta} shows the performance comparison with different $\beta$. We can find that  policies trained using dual gradient descent to adaptively update $\beta$ consistently achieves the best performance overall.\\
\textbf{Sensitivity Analysis (RQ4)}. 
We study how  trade-off hyper-parameters in proposed offline learning methods affect the performance. For  SC, we have trade-off hyper-parameter $\delta$, as shown in Eq.~\ref{Eq:SC},  SC  becomes SDAC when $\delta=0$ or $\delta=1$ . For  PC, if the parameter $\beta \rightarrow \infty $, then  PC resembles the supervised learning. For  DC and RA, the choice of $\alpha$ also creates a trade-off between  supervised learning and our reinforcement learning algorithm. Figure~\ref{fig:sen}  illustrates the  NDCG@10 of SC, PC, DC and RE with different hyper-parameters. The effect of $\beta$ in SR has been studied in RQ3,  and is omitted here. Figure~\ref{fig:sen} shows that we can balance the supervised signal  and task reward 
 by varying these parameters, leading to better  performance under the offline  setting.

\begin{table}
\center
\setlength{\tabcolsep}{1.5mm}{
\resizebox{0.45\textwidth}{!}{
\begin{tabular}{@{}c cccc c cccc c  @{}}
\toprule[1.0pt]
  & \multicolumn{4}{c}{RecSys}  & & \multicolumn{4}{c}{Kaggle}   \\
  \cline{2-5}    \cline{7-10} 
  
     &  H$@$5  & N$@$5  & H$@$10   & N$@$10  & &  H$@$5  & N$@$5  & H$@$10   & N$@$10  &  \\
     \toprule[1.0pt]
   
   SL    &  .3994  & .2824  & .5183   & .3204  & &  .4608  & .3834  & .5107   & .3995  &  \\
       \toprule[1.0pt]
   DQN   &  .3478  & .2417  & .4820   & .2843  & &  .4087  & .3218  & .4524   & .3401  &  \\
    PG   &  .3514  & .2576  & .4883   & .2941  & &  .4172  & .3324  & .4612  & .3517  &  \\
       \toprule[1.0pt]
   SL+DQN    &  .4228  & .3016  & .5333   & .3376  & &  .5069  & .4130  & .5589   & .4289  &  \\
    SL+PG   &  .4325  & .3071  & .5412   & .3414  & &  .5087  & .4172  & .5602  & .4340  &  \\
       \toprule[1.0pt]
   SDAC    &  .3671  & .2624  & .4917   & .3012  & &  .4236  & .3476  & .4721   & .3632  &  \\
    \toprule[1.0pt]
   SC    &  .4216  & .2978  & .5279   & .3351  & &  .4982  & .4052  & .5517   & .4149  &  \\

   SR    &  .4341   & .3086  & .5458   & .3516  & & .5111  & .4239  & .5641  & .4418  &  \\

   PC    &  .4356  & .3074  & .5401   & .3396  & &  .5105  & .4155  & .5627   & .4340  &  \\

   DC    &  $\textbf{.4427}^{\ast}$  & $\textbf{.3219}^{\ast}$  & $\textbf{.5571}^{\ast}$   & $\textbf{.3587}^{\ast}$  & & $ \textbf{.5341}^{\ast}$  & $\textbf{.4339}^{\ast}$  & $\textbf{.5868}^{\ast}$   & $\textbf{.4687}^{\ast}$  &  \\

      RE    &  \underline{.4372}  & \underline{.3102}  & \underline{.5487}  &  \underline{.3527} & &  \underline{.5201}  & \underline{.4278}  & \underline{.5743}   & \underline{.4547}  &  \\
   
  \bottomrule[1.0pt]
\end{tabular}}}
\caption{Purchase performance comparison utilizing  RNN as the backbone. The best and the second best performance  are marked with boldfaces and underlined, respectively.}
\label{Table:sparse}
\end{table}

\section{Conclusions}
In this paper, we presented the first comprehensive analysis of learning interactive recommendation 
offline. We first formalized the interactive recommendation as a probabilistic inference problem,  and then proposed a stochastic and discrete RL algorithm to  maximize user cumulative  rewards. To perform  offline learning effectively, we proposed a general offline learning framework to minimize the distribution mismatch between the  logging policy and learning  policy, including support constraints, supervised regularization, policy constraints, dual constraints and reward extrapolation. We conducted extensive  experiments on two real-world datasets, demonstrating that the proposed  methods can achieve better performance over  existing  methods.

\section*{ Acknowledgments}
The authors would like to thank  the Westlake University and Bright Dream Robotics Joint Institute  for the funding support. The authors also would like to thank  the anonymous reviewers for thoughtful and constructive feedbacks.

\bibliography{references}

\end{document}